\documentclass[11pt,twocolumn]{article}
\usepackage{times}
\usepackage{latexsym}
\usepackage{xcolor,colortbl}
\definecolor{Gray}{gray}{0.85}
\definecolor{darkblue}{rgb}{0, 0, 0.5}
\usepackage{diagbox}
\usepackage{lipsum}
\usepackage{subcaption}
\usepackage{authblk}
\usepackage{graphicx}
\usepackage{hyperref}

\usepackage{natbib}
\renewcommand\cite{\citep}

\hypersetup{colorlinks=true,citecolor=darkblue, linkcolor=darkblue, urlcolor=darkblue}

\newenvironment{conditions}
  {\par\vspace{\abovedisplayskip}\noindent\begin{tabular}{>{$}l<{$} @{${}={}$} l}}
  {\end{tabular}\par\vspace{\belowdisplayskip}}

\author[1]{Florin Brad}
\author[2]{Radu Iacob}
\author[2]{Ionel Hosu}
\author[2]{Traian Rebedea}

\affil[1]{Bitdefender, Romania}
\affil[2]{University Politehnica of Bucharest}

\title{Dataset for a Neural Natural Language Interface for Databases (NNLIDB)}

\date{}

\begin{document}

\maketitle
\begin{abstract}
   Progress in natural language interfaces to databases (NLIDB) has been slow mainly due to linguistic issues (such as language ambiguity) and domain portability. Moreover, the lack of a large corpus to be used as a standard benchmark has made data-driven approaches difficult to develop and compare. In this paper, we revisit the problem of NLIDBs and recast it as a sequence translation problem. To this end, we introduce a large dataset extracted from the Stack Exchange Data Explorer website, which can be used for training neural natural language interfaces for databases. We also report encouraging baseline results on a smaller manually annotated test corpus, obtained using an attention-based sequence-to-sequence neural network. 
\end{abstract}

\section{Introduction}
    Natural language interfaces have gathered a lot of attention as tools for simplifying the interaction between users and computers. These interfaces often exclude or complement input devices, such as keyboard or touch screens, or even specific languages used for interacting with an application. A more focused area is composed of Natural Language Interface to Databases (NLIDB), which would allow a person to retrieve useful information from any database without knowledge of specific query languages such as structured query language (SQL) for relational databases.
    Despite initial efforts into NLIDBs started decades ago, research has advanced slowly and at this moment there are no commercial solutions or widespread prototypes. The main difficulties in solving this problem stem from linguistic failures and the inability to develop general-purpose solutions that are portable to different databases and schemas.
    
    Due to the recent success of deep neural approaches in natural language processing, our aim is twofold. First, we hope to rejuvenate interest in the NLIDB problem by proposing a large dataset, called the Stack Exchange Natural Language Interface to Database (SENLIDB) corpus, for developing data-driven machine learning models and for reporting progress. The training set consists of $24,890$ pairs (textual description, SQL snippet) crawled using the Stack Exchange API that we filtered and cleaned. A smaller test set consisting of $780$ pairs that were manually created by two annotators is also available for comparing solutions.

    Second, we report results on a neural baseline that uses an attention-enhanced sequence-to-sequence (SEQ2SEQ) architecture \cite{Bahdanau2014} to model the conditional probability of an SQL query given a natural language description. This model is trained on the aforementioned dataset and its performance is computed both using cross-validation and on the manually labeled test set. Qualitative results reveal code that is almost syntactically correct and closely related to the user's intention. Moreover, we report results on two smaller tasks, which we call the tables and columns identification tasks. These results suggest that our dataset is indeed valuable for training the first end-to-end neural natural language interface for databases (NNLIDB).
    
    The paper continues with a short overview of related work in natural language interfaces for databases and in similar tasks where deep networks have been successfully employed. Section 3 contains a detailed  description of the large SENLIDB dataset created for training, together with the smaller dataset used for testing and comparing various NLIDB systems. Preliminary results using a SEQ2SEQ neural model with attention trained on the dataset proposed in this paper are presented in Section 4. We then propose alternative indicators for assessing the correctness of generated SQL queries in Section 5, while Section 6 concludes the paper by highlighting the key insights and future work.

\section{Related Work}
As all current NLIDB solutions are using mainly dependency and semantic parsing together with rule-based or constraint-based algorithms, we also present similar problems which inspired our approach, where deep networks have achieved state of the art results. In the last part of the section, we introduce the most frequently used corpora for evaluating the performance of NLIDB systems.

\subsection{Current approaches for NLIDB}
    Natural language interfaces for databases have been studied for decades. Early solutions proposed using dictionaries, grammars and dialogue systems for guiding the user articulate the query in natural language on a step by step basis \cite{Codd74, Hendrix1978}. Most systems developed until mid-90s used a mix of pattern matching, syntactic parsing, semantic grammar systems, and intermediate representation languages for generating the query from text \cite{androutsopoulos1995}. The most important problems encountered by NLIDBs were related to ambiguity in semantics and pragmatics present in natural language: modifier attachment, understanding quantifiers, conjunction and disjunction, nominal compounds, anaphora, and elliptical sentences \cite{androutsopoulos1995}. 
    
    In more recent studies, \citet{Popescu2004} combine syntactic parsing and semantic interpretation for natural language queries to change parse trees such that, by changing the order of some nodes in a tree, it will be correctly interpreted by the semantic analyzer. Then they use a maximum flow algorithm and dictionaries for semantic alignment between the text and several SQL candidates. One of their main contributions is that they introduce a subset of semantically tractable text queries, for which the proposed method generates correct SQL queries in most cases.
    
    NaLIR \cite{NALIR1.Li2014} uses dependency parse trees generated with CoreNLP \cite{CoreNLP:2014} and several heuristics and rules to generate mappings from natural language to candidate SQL queries. Given the dependency tree, the database schema and associated semantic mappings, the system proceeds in building alternative query trees which can be easily translated to SQL. To determine the best query tree, the system combines a scoring mechanism and an interaction with the user to select the best choice (from a list of reformulations of the query tree into natural language). The scoring for each query tree takes into account the number of alterations performed on the dependency tree in order to generate it, the database similarity/proximity between nodes adjacent in the query tree, and the syntactic correctness of the generated SQL query.
    
    The most promising results reported on several databases used for validating NLIDBs have been recently achieved by Sqlizer \cite{Yaghmazadeh0DD17}. Its main contributions are related to the fact that it uses a semantic parser to generate a query sketch, which is then completed using a rule based system, and iteratively refined and repaired using rules and heuristics until the score of the generated SQL query cannot be improved. Sqlizer is one of the few systems which employs machine learning and Word2Vec \cite{Mikolov:2013} for generating the query sketch - a general form of the query, including clauses, but which does not contain any specific database schema information (e.g. table and column names). 
    
\subsection{Deep learning solutions for related problems}
    To the best of our knowledge, no deep learning solution has been proposed for the NLIDB problem until now, mainly due to the lack of large datasets for training such complex models. However, neural models have been successfully used for similar problems.

    \citet{Mou2015} introduced a case study for code generation from problem descriptions using recurrent neural networks (RNN). They trained a SEQ2SEQ architecture with a character-level decoder and produced program snippets that are almost executable and retain functionality. Moreover, they showed that the RNN generates novel code alternatives compared to the programs seen during training, thus ruling out the possibility that the network merely memorizes the input examples. \citet{Ling2016} combined the SEQ2SEQ approach with a pointing mechanism \cite{Vinyals2015} in order to generate Python and Java code using textual descriptions automatically extracted from collectible trading card games. 
    
    More recently, \citet{Yin2017} proposed a syntax-aware neural model that generates Abstract Syntax Trees from natural language descriptions, which then get mapped deterministically to the target source code. The decoder is guided by a predefined grammar, so their solution is agnostic of the target programming language. Using this syntax aware decoding mechanism, they show to improve the SEQ2SEQ baseline for code generation. 
    
    Another related topic is semantic parsing using deep neural networks. Semantic parsing focuses on converting natural language into logical forms which are used for querying knowledge bases \cite{Berant13} and has also been successfully used for NLIDBs. Recent neural approaches for semantic parsing \citep{Dong16, Herzig17} use a SEQ2SEQ network that maps natural language text to logical forms. Recent solutions bypass the need for ground truth logical forms and instead train a supervised neural model from query-answer pairs \citep{yin2015, neelakantan17}.

\subsection{Existing corpora for NLIDB evaluation}
\begin{table*}
\begin{center}
\begin{tabular}{|l|r|r|r|r|}

\hline
    \textbf{Dataset} & \textbf{\# Tables} & \textbf{\# Columns} & \textbf{\# Text queries} & \textbf{\# SQL queries} \\ \hline
    ATIS & 27 & - & 2,866 & N/A \\ 
    NLmaps & N/A & N/A & 2,380 & N/A \\ 
    MAS & 17 & 53 & 196 & 196\\ 
    IMDB & 16 & 65 & 131 & 131\\
    Yelp & 7 & 38 & 128 & 128\\
    SENLIDB Train & 29 & 204 & 24,890 & 24,890 \\ 
    SENLIDB Test & 15 & 98 & 780 & 296\\  \hline
    
\end{tabular}
\end{center}
\caption{\label{table:Datasets} Comparison of existing datasets and the SENLIDB corpora for NLIDB systems}
\end{table*}

    Solutions to the NLIDB problem have been traditionally evaluated against databases with few tables and on validation datasets with a small number of entries. 
    
    One of the most complex databases for NLIDB evaluation is ATIS (Air Travel Information Corpus) \cite{ATIS}, which stores information about data flights and features 27 tables. However, it only has 2,886 natural language queries and no corresponding SQL statements, making it unsuitable for a data-driven approach. Most recent systems have moved to validation datasets which contain both the natural language query and the corresponding SQL snippet, such as MAS (Microsoft Academic Search), IMDB, and Yelp. For example, Sqlizer \cite{Yaghmazadeh0DD17} achieves 80\% accuracy on MAS, while NaLIR \cite{NALIR1.Li2014} obtains only 32\% accuracy on the same data. There also exist some slightly larger corpora for querying geolocation databases, the largest being NLmaps \cite{haas2016corpus} which contains 2,380 text queries but with no corresponding SQL code (instead they use machine readable language - MRL for expressing queries).
    
    The training set (SENLIDB Train) proposed in this paper is by far larger than any of the existing datasets, as can be seen from Table 1. This makes it extremely useful for training solutions using machine learning, including neural NLIDBs. More, the test set (SENLIDB Test), which has been manually annotated by two experts, is twice as large as current validation corpora and contains several  text formulations for the same SQL query.

\section{Dataset construction}

A deep neural architecture, such as SEQ2SEQ, requires a large number of input-output pairs to produce qualitative results. The next subsections describe the steps taken to build the SENLIDB dataset, including our attempts to correct some of the problems inherent with crowdsourced data.

\subsection{Data crawling and preprocessing}
    The Stack Exchange Data Explorer allows users to query the entire database of the well-known question-answering platform through a public API\footnote{\url{http://data.stackexchange.com/stackoverflow/query/new}}. 
    The database uses Microsoft SQL Server, therefore users query it using the SQL extension developed by Microsoft, called Transact-SQL (T-SQL). For each query to the Stack Exchange database issued by a user, the web interface enforces the user to add a title and also an optional longer description. The main rationale for these two fields is for users to provide an accurate textual description for each query they make. However, there is no method to ensure that the title or the description entered for a query are actually relevant in describing it. 
    
    The list of all user queries is available online\footnote{\url{http://data.stackexchange.com/stackoverflow/queries}} and Stack Exchange offers various sorting and filtering capabilities including most upvoted or viewed queries. An important characteristic is that all available queries are correct, meaning that they do not throw any errors when querying the Stack Exchange database. Moreover, some of them are "interactive" - users can input values in the web interface for temporary variables enclosed by '\#\#' or '\#' in the SQL query.
        
    In order to build the proposed dataset, we started by crawling all user queries from Stack Exchange, as they appear in the section 'Everything' ordered descending by creation date \footnote{\url{http://data.stackexchange.com/stackoverflow/queries?order\_by=everything}}.
    First of all, we discarded SQL snippets longer than $2,000$ characters as we considered them to be too complex. This step resulted in about $2,000,000$ queries. The next step was to create pairs of textual description (which included the title and the actual description of a query) and corresponding SQL snippet. We then removed duplicate pairs (identical SQL code and description) and approximately 600,000 pairs were left. After this step, we removed items with SQL code in the description using simple empirical rules (descriptions starting with 'select' and containing 'from'). The remaining dataset was reduced to roughly 170,000 pairs. 
    
    Afterwards, we removed the comments from the SQL snippets and eliminated the entries that now had void snippets. Finally, we took away items with identical textual descriptions and different SQL snippets. For description $d$ and corresponding SQL snippets $s_1, ..., s_n$, we kept the code snippet $s_i$ of median length, as we consider that an average length description is probably better than very long and very short ones which are probably outliers.
    This resulted in a dataset with 24,890 items, each having an unique textual description and an associated SQL query. 
    
    Although descriptions in this dataset are unique, there are $2,225$ identical SQL queries with different descriptions.
     
\subsection{Large dataset for training and validation}
    
\begin{table*}[t]

  \centering
  
  \begin{tabular}{|l|r|r|r|r|r|r|}
  \hline 
  \backslashbox{\textbf{\# SQL query tokens}}{\textbf{\# text tokens}} & \textbf{{1-10}} & \textbf{{11-25}} & \textbf{{26-50}} & \textbf{51-100} & \textbf{100+} & \textbf{Total} \\ \hline
  \textbf{2-4}   & 2094 & 3321 & 2634 & 1536 & 605  & 10190\\ \hline
  \textbf{5-10}  & 641  & 2547 & 3182 & 2306 & 742  & 9418 \\ \hline
  \textbf{11-20} & 121  & 724  & 1150 & 876  & 318  & 3189 \\ \hline
  \textbf{21-50} & 21   & 239  & 470  & 584  & 266  & 1580 \\ \hline
  \textbf{51+}   & 1    & 10   & 35   & 99   & 72   & 217 \\ \hline
  \textbf{Total}   & 2878 & 6841 & 7471 & 5401 & 2003 & \\ \hline
  \end{tabular}
  \caption*{(a) Length statistics for the training dataset}
  \label{tab:SQLDescr_Train}

\bigskip

  \begin{tabular}{|l|r|r|r|r|r|r|}
  \hline
  \backslashbox{\textbf{\#SQL query tokens}}{\textbf{\# text tokens}} & \textbf{{1-10}} & \textbf{{11-25}} & \textbf{{26-50}} & \textbf{{51-100}} & \textbf{Total} \\ \hline
  \textbf{{2-4}}   & 88   & 1    & 0    & 0    & 89\\ \hline
  \textbf{{5-10}}  & 270  & 69   & 8    & 4    & 351 \\ \hline
  \textbf{{11-20}} & 77   & 181  & 23   & 4    & 285 \\ \hline
  \textbf{{21-50}} & 1    & 34   & 18   & 2    & 55 \\ \hline
  \textbf{Total}   & 436  & 285  & 49   & 10   & \\ \hline
  \end{tabular}
  \caption*{(b) Length statistics for the test dataset}
  \label{tab:SQLDescr_Test}

  \caption{Overview of the number of tokens from the SQL snippet and the textual description for the SENLIDB corpora}
  \label{table:SQLDescr}
  
\end{table*}
    
    We consider that the previously described dataset can be used effectively for training machine learning models for NLIDB, including more data-hungry models such as neural NLIDBs. As this corpus was created by a large number of users from the Stack Exchange data portal, one might expect that the quality of the entries to be similar to other corpora created using various crowdsourcing mechanisms. To this extent, although this dataset can also be used for validation (using either cross-validation or a hold-out set), the results will be impacted by the inherent biases, noise and errors collected through crowdsourcing. Some of the particularities of these data are addressed next.
    
    First, most of the SQL snippets are relatively simple, containing at most 10 distinct tokens, as can be easily seen in Table 2.
    In contrast, textual descriptions are more evenly distributed, based on the number of tokens, with $2,003$ of the entries in the dataset having more than $100$ tokens. Thus although some queries might have an incomplete textual description, most of them are well explained.

\begin{table}
\begin{center}
\begin{tabular}{|l|r|r|}

\hline
    \textbf{Table name} & \textbf{\# occur. train} & \textbf{\# occur. test} \\ \hline
    Posts & 15159 & 383 \\ 
    Users & 7672 & 229\\ 
    Tags & 4765 & 134\\ 
    Posttags & 3370 & 39\\ 
    Votes & 2476 & 22\\ 
    Comments & 1583 & 41\\ 
    Posthistory & 1214 & 2 \\ 
    Badges & 625 & 16\\ 
    Posttypes & 616 & 4\\ 
    Votetypes & 336 & 6\\
    Other tables & 1080 & 16\\ \hline
    
\end{tabular}
\end{center}
\caption{\label{table:TableFreq} Most frequent table names in SENLIDB sorted descending by occurrences in training set}
\end{table}

\begin{table}
\begin{center}
\begin{tabular}{|l|r|r|}

\hline
    \textbf{SQL expr.} & \textbf{\# occur. train} & \textbf{\# occur. test} \\ \hline
    select & 22145 & 295 \\ 
    from & 21982 & 295 \\ 
    where & 18894 & 203\\ 
    order & 13114 & 77 \\ 
    count & 8294 & 57 \\
    join  & 7943 & 29 \\ 
    group & 7366 & 27 \\ \hline
    
\end{tabular}
\end{center}
\caption{\label{table:Expr} Most frequent SQL expressions in \\SENLIDB}
\end{table} 

    Second, the Stack Exchange database schema available in the dataset contains 29 tables. Interestingly, their actual appearances in the dataset, judging by the number of occurrences in individual queries, follows Zipf's law \cite{zipf2016human} as it can be observed in Table 3. We note that a large majority of queries refer to the 'Posts' and 'Users' tables, while other tables make almost no appearance in the dataset (e.g. 'PostNotices', 'PostNoticeTypes'). In Table 4 we present the most frequent SQL expressions in the datasets. Half of the queries contain ordering clauses and almost a third include multiple joined tables and group by clauses.     
    
    Third, the dataset contains samples of varied difficulty, from simple select operations to complex nested queries. We computed the Halstead complexity metrics \cite{halstead1977elements} to gain an insight into the difficulty of the SQL snippets in our datasets. To measure the difficulty of a snippet we used the formula \cite{halstead1977elements}: 
    \begin{equation} \label{eq:1}
    Difficulty = \frac{\eta_1}{2} \cdot \frac{N}{\eta_2}
    \end{equation}
    where:
    \begin{conditions}
     \eta$$_1$$ & number of distinct operators \\
     \eta$$_2$$ & number of distinct operands \\   
     N      & total number of operands
    \end{conditions}

\begin{figure}

\begin{subfigure}{\linewidth}
   \includegraphics[width=1\linewidth]{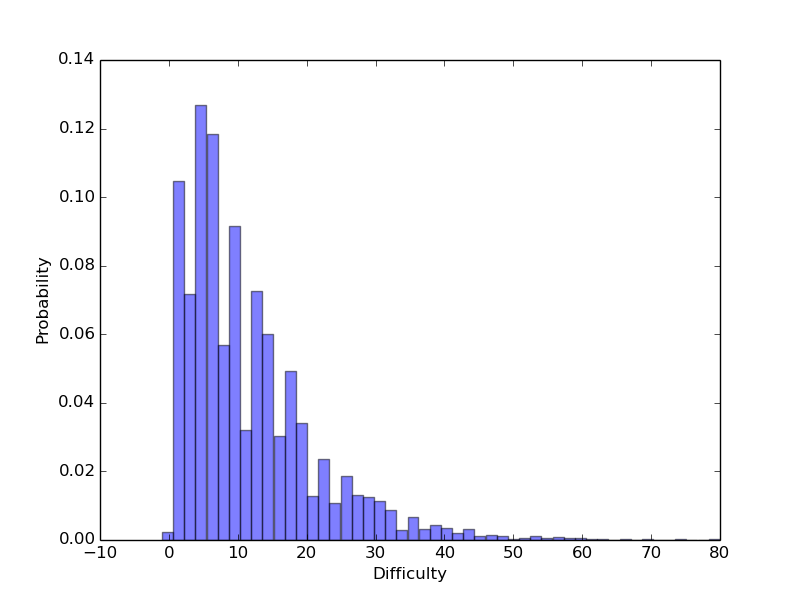}
   \caption{Training dataset difficulty}
   \label{fig:DiffTrain} 
\end{subfigure}

\begin{subfigure}{\linewidth}
   \includegraphics[width=1\linewidth]{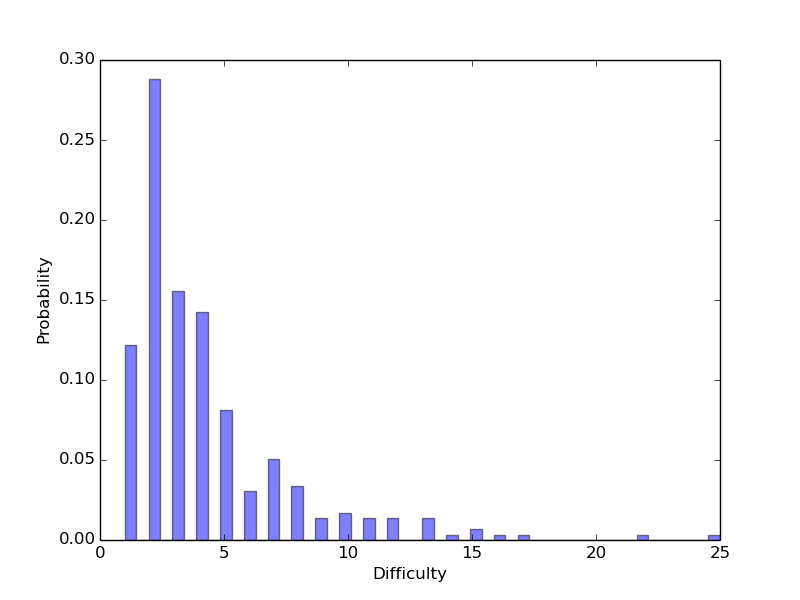}
   \caption{Test dataset difficulty}
   \label{fig:DiffTest}
\end{subfigure}

\label{fig:Train_Difficulty_Histo}
\caption{Histograms of the Halstead difficulty measure for the training (a) and  test (b) sets}

\end{figure}

Finally, we used an off-the-shelf library \footnote{\url{https://pypi.python.org/pypi/polyglot}} to detect the language of the query descriptions. More than 95\% were classified as English, followed at a great distance by French and Russian with less than 100 entries each. We remarked that some of the descriptions contain table and column names, which could affect the language identification performance (with a small bias towards English). 

\subsection{Manually annotated test dataset}
    In order to have a reliable test and validation dataset for the Stack Exchange database, we also developed a smaller corpus which was manually annotated by two senior undergraduate students in Computer Science. The SQL queries included in the test dataset are a subset of the data collected from the Stack Exchange Data Explorer as previously described. Each query has been labelled by at least one annotator using between 1 and 3 different textual descriptions that describe the respective SQL snippet in natural language (English). The annotators then ran the query in the interface and verified that the returned results are correct and correspond to the description. The total number of distinct queries is 296, while the number of textual annotations is 780, averaging to $2.63$ textual reformulations per query. 

    In order to facilitate the annotation process, the annotators used an application which allowed the user to view a SQL query from the original dataset and add one or more possible descriptions. The SQL queries chosen for manual annotation were randomly selected from those with a very short textual description in the original corpus, consisting of only 1-2 tokens. These items were considered not informative enough to be included in the training set and were thus added to the human-annotated test set.
    
    In order to achieve a better understanding of how similar or different the produced annotations are, for each sample we computed the BLEU score \cite{papineni2002bleu}, with the smoothing function proposed in \citet{chen2014systematic}, between the descriptions of one annotator and those produced by the other annotator. The average of the scores obtained for each sample was $57.10$, which is consistent with inter-translator BLEU scores observed in machine translation.

    It is important to notice that there are some differences between the train and test dataset. The most important one is highlighted in Figure 1 where the Halstead difficulty score for the test set is concentrated between 0-5 as opposed to the train dataset where the mode is at 8. This means that the queries in the test dataset are slightly less complex than the queries in the train dataset. There are also some other differences between the two datasets, such as the distribution of query and description sizes (see Table 2) and most frequent table names (see Table 3); these arise from how the test SQL queries were subsampled and annotated.

\section{Model and experiments}

For what we consider to be the first end-to-end neural NLIDB, we trained a SEQ2SEQ model with attention on the (description, SQL) pairs in the SENLIDB train set. We used the open-source neural machine translation toolkit OpenNMT (\url{http://opennmt.net/}) \cite{DBLP:journals/corr/KleinKDSR17}. This system implements a standard SEQ2SEQ model with global-general-attention \cite{luong2015effective}. Both the encoder and the decoder are long short-term memory (LSTM) cells with two hidden layers and 500 neurons. The word embedding layer has 500 neurons. We used batches of maximum size 64. We trained the models with Stochastic Gradient Descent (SGD) for 25 epochs with a learning rate of 1.0 and a learning decay of 0.5 if perplexity did not decrease on the validation set. 
We generated SQL statement using a beam search of size 5.

\begin{table}
    \centering
    \begin{tabular}{|l|r|r|}
    \hline
    \textbf{Dataset} & \textbf{Perplexity} & \textbf{BLEU} \\
    \hline
    Validation & 1.16 & 16.9 \\
    Test-original & 1.24 & 13.5 \\
    Test-annotated & 1.23 & 18.2 \\
    \hline
    \end{tabular}

    \caption{\label{table:testResults} Perplexity and BLEU score of the generated SQL statements on the validation set, original test set and annotated test set descriptions}
\end{table}

\begin{table*}[t]
\begin{center}
\begin{tabular}{|p{7.5cm}|p{7.5cm}|}

\hline
    \textbf{Natural language description} & \textbf{Predicted SQL query} \\ \hline
    top 50 users from india & select top 50 id , displayname , reputation , websiteurl , location from users where location like india order by reputation desc \\ \hline
    top 100 users with no of gold badge & select top 100 b . userid , u . displayname , count ( * ) as gold from badges b inner join users u on u . id = b . userid group by b . userid , u . displayname order by count ( b . name ) desc \\ \hline
    user rep & select reputation , count ( * ) from users group by reputation order by reputation \\ \hline
    text of highest - scored comment & select top 100 id as [ comment link ] , score , text from comments order by score desc \\ \hline
    questions with most view & select top 500 id as [ post link ] , viewcount from posts order by viewcount desc \\ \hline    
\end{tabular}
\end{center}
\caption{\label{table:SQLExamples} Examples of SQL queries generated by the SEQ2SEQ model with attention}
\end{table*}

\begin{table}
    \centering
    \begin{tabular}{|l|r|r|r|}
    \hline
    \textbf{Dataset} & \textbf{Precision} & \textbf{Recall} & \textbf{F1 score} \\
    \hline
    Validation & 0.71 & 0.55 & 0.62 \\
    Test-original  & 0.51 & 0.41 & 0.45 \\
    Test-annotated  & 0.82 & 0.72 & 0.76 \\
    \hline
    \end{tabular}

    \caption{\label{table:entityTaskResults} Precision, recall and F1 score for the tables identification task}
\end{table}

\begin{table}
    \centering
    \begin{tabular}{|l|r|r|r|}
    \hline
    \textbf{Dataset} & \textbf{Precision} & \textbf{Recall} & \textbf{F1 score} \\
    \hline
    Validation & 0.65 & 0.47 & 0.54 \\
    Test-original & 0.35 & 0.29 & 0.31 \\
    Test-annotated  & 0.55 & 0.47 & 0.50 \\
    \hline
    \end{tabular}

    \caption{\label{table:attributeTaskResults} Precision, recall and F1 score for the columns identification task}
\end{table}

Similarly to \citet{Ling2016}, we report the BLEU score between the generated SQL queries and the ground truth SQL snippets in Table 5. While the BLEU score could penalize differently written, but otherwise correct, code snippets, it is still useful to measure the degree of token overlap. The results are reported for a validation set (holdout of $4,000$ random samples from the train set) and for the test set, using both the original and the manually annotated texts.
We notice similar perplexities for SQL code generated from the original test titles and from the manually annotated ones, which means that both generate likely code. This is to be expected as the decoder is trained on SQL select statement therefore it will probably generate some sort of select statement even for short input texts given to the encoder. However, the original titles are much shorter compared to the annotated titles, and so the more informative natural language descriptions yield a SQL query that resembles more closely the ground truth SQL under a BLEU score. Thus, although both shorter (incomplete) and longer (and more descriptive) texts generate likely SQL statements, the more descriptive manually annotated texts generate queries significantly more similar to the ground truth (BLEU score $18.2$ vs $13.5$, as reported in Table 5). 

The initial vocabulary for the encoder (text descriptions) had $6,000$ tokens, while the vocabulary of the decoder (SQL queries) consisted of $16,000$ tokens. This resulted in a very large embedding matrix, thus we decided to restrict the number of tokens for both encoder and decoder to $500$ and $2,000$, respectively, by keeping only the most frequent tokens and replacing the others with the UNK token. Reducing the size of the vocabularies for both encoder and decoder resulted in a significant improvement for the performance of the model (BLEU score $18.2$ vs $13.06$ for the annotated test set). 

From a qualitative perspective, Table 6 provides several examples of SQL queries generated for the validation set. The generated SQL statement are syntactically correct most of the time even when the textual description is incomplete or use abbreviations (e.g. "no" for "number). More, in the second example, we can also observe that the model learns to use table aliases correctly in complex queries with joined tables. On another hand, although the generated queries are syntactically correct, in most cases they fail to return the desired results when they are executed against the database. When the system fails to generate the correct SQL query for a  description, it still generates a query related to the natural language description. 

It is important to mention that, in order to correctly write an SQL statement, one needs to know the schema of the database. This is an aspect that we did not take into consideration when training the baseline model. Thus the model is not explicitly provided with the database schema, however it can infer it from the training set. However, we believe that more complex approaches that integrate schema information and are syntax-aware can produce better results than a SEQ2SEQ model.

\section{Discussion}

Given that, unlike natural language, SQL is highly restricted and unambiguous, we believe that the problem of generating SQL queries from natural language can be reduced to a number of independent sub-problems. For example, in order to retrieve the desired information from a database, the appropriate table columns need to be instantiated in the SELECT clauses, and the correct tables need to be instantiated in the FROM clause. Breaking down the complex task of automatically generating SQL in multiple simpler tasks and working on each task separately can, in our opinion, yield significant improvements faster.

Apart from the BLEU score, we propose two new tasks that are easier than the NLIDB problem. This approach stemmed from the difficulty of the problem and the need for a more structured grasp of the performance of a certain system on this task. Therefore, we chose to also evaluate the ability of the proposed NNLIDB to correctly instantiate tables and columns from the database schema. For these two tasks, the most important metrics are precision and recall. For example, given a sample from the dataset, we compare the SQL query generated by the neural network architecture with the correct SQL statement and count existing and missing table and column names.

In tables 7 and 8 we evaluate the performance of our baseline on the tables and columns identification tasks. We observe that on the validation and annotated test set, precision and recall scores are significantly higher, due to the fact that these are more informative than the original test set descriptions. Given the fact that the database schema contains a total of 29 entities (table names) and 204 attributes (column names), the precision and recall scores prove that the baseline model delivers decent performance on these tasks and moreover, that both tasks are representative for measuring the performance of a system on the NLIDB problem. It is important to mention that, for the sake of simplicity, for the columns identification task we ignored the fact that in different tables there may be columns with the same name (e.g. "id").

Both the tables and columns identification tasks can be made more difficult using stricter evaluation. For example, for the table task, one could consider only the entities that are instantiated strictly in the FROM clause and the attributes that are instantiated in the SELECT clause.

\section{Conclusions}
In this paper we have introduced new datasets for training and validating natural language interfaces to databases. The SENLIDB train dataset is the first large corpus designed to develop data-driven NLIDB systems and it has been successfully used to train an end-to-end neural NLIDB (NNLIDB) using a SEQ2SEQ model with attention. Although the generated SQL output may sometimes be syntactically invalid and is rarely the desired SQL statement for the given textual query, the results are promising.

The pursuit of a successful NNLIDB is still at the beginning and we hope that the current research will provide the first steps needed to investigate more complex solutions. Future research will investigate whether using a stacked decoder - one for generating a query sketch (e.g. subclauses) and one for the elements related to the database schema - will provide a better solution. 

In comparison with existing approaches for NLIDB systems, our solution does not use any rules, heuristics or information about the underlying database schema or SQL syntax. On the other hand, the generated SQL queries are more often than not inaccurate and thus we have not compared the accuracy of the NNLIDB with existing solutions. However, we have focused on verifying how similar the generated SQL queries are to the annotated ones using measures from machine translation (BLEU) and also precision and recall for simpler tasks, such as generating the correct table and column names in a SQL statement. 

\bibliography{main}
\bibliographystyle{plainnat}

\end{document}